\documentclass[conference]{IEEEtran}
\IEEEoverridecommandlockouts

\usepackage[utf8]{inputenc}
\usepackage[T1]{fontenc}
\usepackage{url}
\usepackage{hyperref}
\usepackage{cite}
\usepackage{graphicx}
\usepackage{booktabs}
\usepackage{textcomp}
\usepackage{amsmath}
\usepackage{enumitem}

\usepackage{booktabs}
\title{From Players to Champions: A Generalizable Machine Learning Approach for Match Outcome Prediction with Insights from the FIFA World Cup\\
}

\author{\IEEEauthorblockN{Ali Al-Bustami\textsuperscript{*}}
\IEEEauthorblockA{\textit{Department of Electrical and Computer Engineering} \\
\textit{University of Michigan-Dearborn}\\
Dearborn, MI 48128, USA \\
abustami@umich.edu \\
\textsuperscript{*}Corresponding author}
\and
\IEEEauthorblockN{Zaid Ghazal}
\IEEEauthorblockA{\textit{Department of Computer and Information Science} \\
\textit{University of Michigan-Dearborn}\\
Dearborn, MI 48128, USA \\
zghazal@umich.edu}
}

\begin{document}

\maketitle
\thispagestyle{empty}
\pagestyle{empty}

\begin{abstract}
Accurate prediction of FIFA World Cup match outcomes holds significant value for analysts, coaches, bettors, and fans. This paper presents a machine learning framework specifically designed to forecast match winners in FIFA World Cup. By integrating both team-level historical data and player-specific performance metrics such as goals, assists, passing accuracy, and tackles, we capture nuanced interactions often overlooked by traditional aggregate models. Our methodology processes multi-year data to create year-specific team profiles that account for evolving rosters and player development. We employ classification techniques complemented by dimensionality reduction and hyperparameter optimization, to yield robust predictive models. Experimental results on data from the FIFA 2022 World Cup demonstrate our approach’s superior accuracy compared to baseline method. Our findings highlight the importance of incorporating individual player attributes and team-level composition to enhance predictive performance, offering new insights into player synergy, strategic match-ups, and tournament progression scenarios. This work underscores the transformative potential of rich, player-centric data in sports analytics, setting a foundation for future exploration of advanced learning architectures such as graph neural networks to model complex team interactions.
\end{abstract}

\section{Introduction}
Football, or soccer, is one of the most popular sports globally, with the FIFA World Cup being its most prestigious tournament. The ability to accurately predict match outcomes is of great interest to fans, coaches, and bettors, offering insights for strategic planning and decision-making. With the rise of Machine Learning (ML), sports analytics has seen significant advancements, enabling new ways to analyze and forecast match outcomes based on historical data.

\begin{figure}
    \centering
    \includegraphics[width=0.99\linewidth]{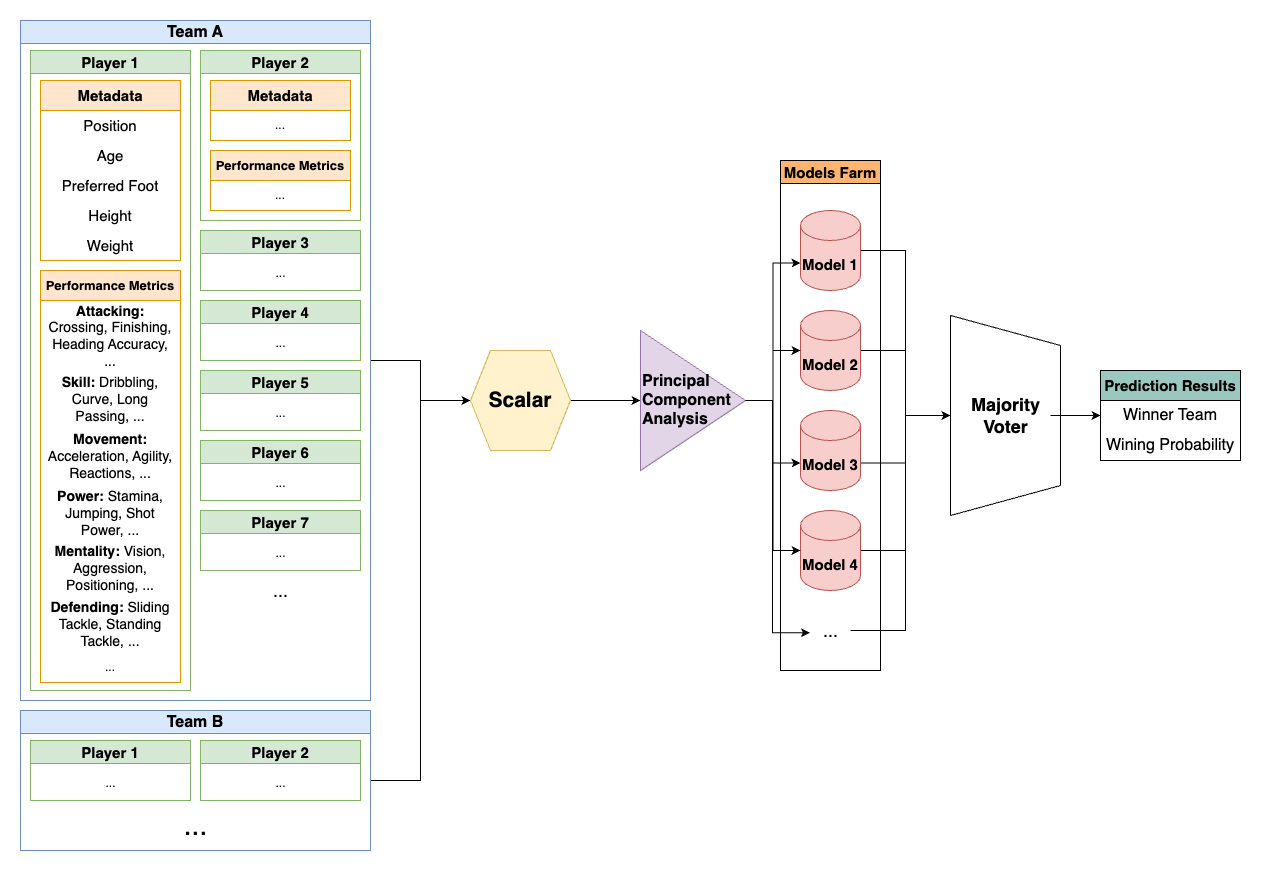}
    \caption{Machine Learning (ML) Framework for Match Outcome Prediction
This diagram illustrates a predictive pipeline that processes player metadata and performance metrics, applies feature scaling and Principal Component Analysis (PCA), and utilizes an ensemble of models. A majority voting mechanism aggregates predictions to determine the winning team and its probability.}
    \label{fig:ml_framework}
\end{figure}

Existing research has applied various ML techniques to predict football match outcomes, primarily focusing on team-level statistics such as historical match results and goal differences. These studies have utilized models like logistic regression, decision trees, random forests, and neural networks \cite{RODRIGUES2022463}\cite{AttaMills2024}. Some studies have incorporated player-specific data to capture individual contributions \cite{WONG2025100537}. However, most of these studies are focused on league matches, which may not fully capture the unique dynamics of the World Cup. The World Cup's unique structure, with group stages and knockouts, introduces additional strategic considerations that may not be present in league matches. Furthermore, national teams assemble players from different clubs, making player interaction data less consistent compared to club teams. There is a lack of research that effectively models team dynamics using detailed player interactions, particularly in the World Cup context.

To address these limitations, this paper proposes new ML-based framework, shown in Figure~\ref{fig:ml_framework} , designed for predicting soccer match winner by integrating player-level and team-level data, applying feature scaling and dimensionality reduction, and employing an ensemble of classifiers to determine the winning team. This design leverages both individual player attributes and collective team performance indicators. To evaluate, we conducted benchmarking using data from the FIFA 2022 World Cup and a baseline model, comparing the framework predictions with the actual match results and the baseline model predictions.

\section{Related Works}
The application of ML to predict football match outcomes has gained significant attention in recent years, driven by the availability of historical data and advancements in computational techniques. Existing studies have primarily focused on team-level statistics, such as historical match results, goal differences, and FIFA rankings, to forecast outcomes. For instance, Rodrigues and Dias employed logistic regression and decision trees to predict football match results, achieving encouraging performance in betting profit margins by leveraging team performance metrics \cite{RODRIGUES2022463}. Similarly, Atta Mills et al. explored random forests and neural networks to forecast soccer match outcomes in the English Premier League, demonstrating improved accuracy over traditional statistical methods when using aggregated team data \cite{AttaMills2024}. Zeileis and Hornik utilized a random forest model blending historical match data, team rankings, and covariates like market value to predict the 2022 FIFA World Cup, providing probabilistic forecasts for all possible matches \cite{zeileis2022machine}.

Beyond team-level approaches, some researchers have recognized the value of player-specific data in enhancing prediction accuracy. Wong et al. developed a model incorporating individual player statistics, such as goals, assists, and passing accuracy, to predict soccer outcomes, showing that player attributes can capture nuanced contributions overlooked by team-only models \cite{WONG2025100537}. Saha et al. proposed a deep learning framework using deep neural networks and artificial neural networks for football match prediction, achieving 63.3\% accuracy on 2018 World Cup matches, though their dataset primarily relied on rankings and team performance rather than detailed player statistics \cite{saha2020deep}. These studies, while insightful, predominantly focus on league matches, where player interactions are more consistent due to stable team rosters across a season.

In the context of the FIFA World Cup, predictive modeling faces unique challenges due to its tournament structure, which includes group stages and knockout rounds, and the assembly of national teams from diverse club backgrounds. Efforts to predict World Cup outcomes have been limited but notable. For example, TGM Research applied ML techniques, including support vector machines and neural networks, to forecast the 2022 FIFA World Cup results, relying heavily on team rankings and historical tournament data \cite{tgm2022predicting}. While achieving reasonable accuracy, their approach did not account for player interactions or the dynamic shifts in team performance during the tournament. Similarly, a mathematical model by Oxford researchers, based on probabilistic methods and team strength estimates, provided a benchmark for World Cup predictions but lacked the granularity of player-level data \cite{bull2022mathematician}.

Graph-based methods have emerged as a promising direction in sports analytics, though their application to football remains underexplored. Huang et al. utilized graph neural networks (GNNs) to predict outcomes in American football and esports, modeling player interactions as graphs to capture team dynamics \cite{huang2021graph}. Their findings suggest that GNNs can outperform traditional ML models by learning complex relationships within teams, a concept yet to be fully applied to World Cup football predictions. Recent work by [Author TBD] further explored GNNs for predicting football formations, highlighting their potential to analyze player interactions and optimize team strategies, though the focus was not on match outcomes \cite{ai2024graph}.

Despite these advancements, significant gaps persist. Most existing models are tailored for league competitions, where team compositions remain relatively stable, and do not fully address the World Cup’s unique dynamics, such as the impact of tournament progression or the variability in national team player interactions. The reliance on team-level statistics often overlooks the synergistic effects of individual players, which are critical in high-stakes matches. Furthermore, while player-specific data has been incorporated in some studies, there is a lack of research effectively modeling team dynamics using detailed player interactions in the World Cup context. This limitation is compounded by the scarcity of approaches that adapt advanced techniques like GNNs to football, particularly for tournament settings.

This paper addresses these gaps by introducing an ML framework specifically designed for FIFA World Cup match prediction. Our approach integrates team-level historical data with detailed player-specific performance metrics, such as goals, assists, passing accuracy, and tackles, to model the complex interactions within teams. Unlike previous models that rely solely on aggregate statistics, our framework captures the evolving dynamics of national teams and their players over time. By leveraging advanced techniques such as dimensionality reduction, hyperparameter optimization, and majority voting, we ensure the robustness and generalizability of our predictions.

\section{Methodology}
\label{sec:methodology}
\begin{figure*}[t]
    \centering
    \includegraphics[width=0.95\textwidth]{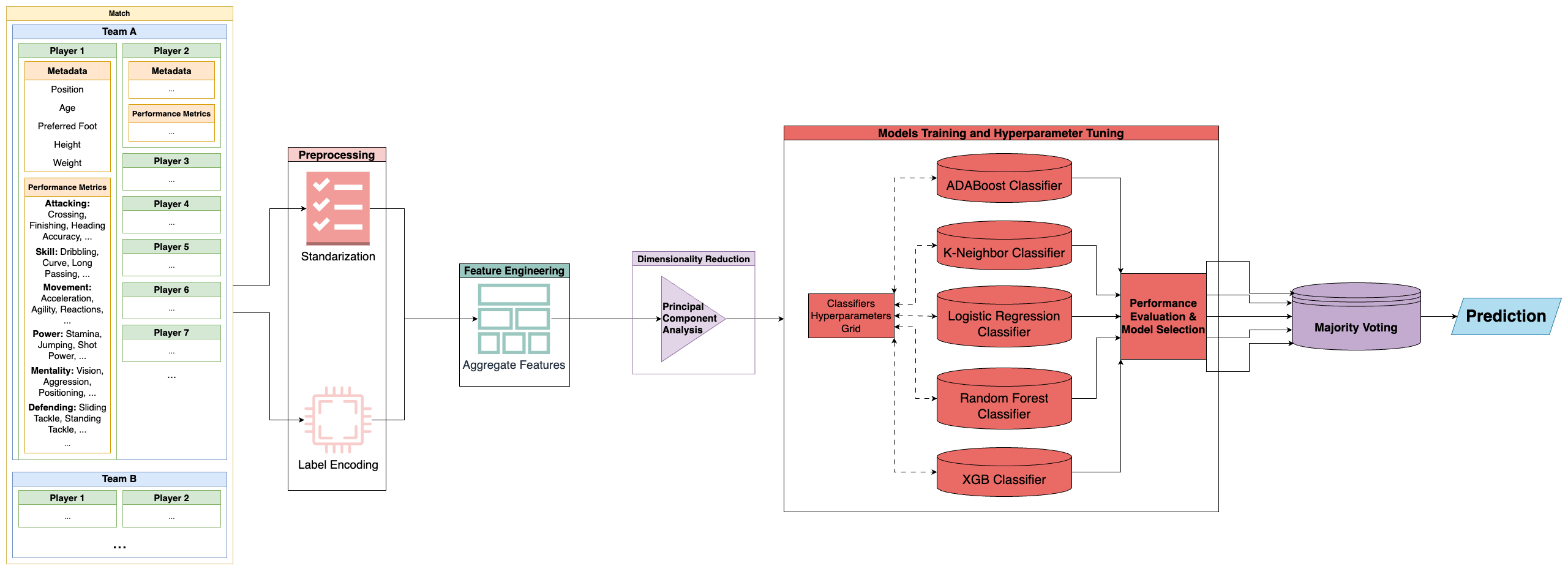}
    \caption{Proposed training pipeline for FIFA World Cup outcome prediction.
    The pipeline includes data collection from player‐level and match‐level attributes, label encoding, preprocessing, feature engineering, dimensionality reduction, cross‐validation‐based hyperparameter tuning, performance evaluation and model selection, and final majority‐voting ensembling to predict match outcomes.}
    \label{fig:training_pipeline}
\end{figure*}

This section outlines the framework training and inference mechanisms for forecasting match outcomes using team-level representations derived from player performance attributes. By leveraging insights from sports analytics and integrating ML algorithms, we construct predictive models that estimate each team's probability of victory based on multiple data sources. Figure~\ref{fig:training_pipeline} illustrates the training process. It begins with data collection from both player-level and match-level attributes, followed by label encoding, feature engineering, dimensionality reduction, and cross-validation-based model selection. We then use majority voting among multiple classifiers to generate a single outcome prediction.

\subsection{Data Assembly and Feature Extraction}

\subsubsection{Data Sources}  
We collected two primary categories of data. The first category, Player-Level Data, consists of individual player attributes gathered from national team rosters spanning multiple years (2015--2022, along with a partial 2023 dataset). These attributes capture technical skills, physical capabilities, mental competencies, and goalkeeping proficiencies, aligning with existing research that highlights their importance in assessing player and team performance~\cite{baboota2019predictive,berrar2019incorporating,constantinou2013profiting}. The second category, Match Outcomes, includes historical match results with details such as home and away teams, final score, net score, and tournament context (e.g., World Cup, continental championships, friendlies)~\cite{hucaljuk2011predicting}. 

Prior research emphasizes the significance of various player-specific metrics in shaping match dynamics~\cite{berrar2019incorporating}. In accordance with these findings, we prioritized technical statistics such as passing, crossing, and dribbling, along with defensive and goalkeeping indicators, ensuring that the extracted data comprehensively captures relevant aspects of player and team performance.

\subsubsection{Feature Selection}  
To align individual attributes with team-level performance, we followed a structured selection process based on three key principles. The first principle, relevance, ensures that selected features directly quantify a player's offensive, defensive, or overall proficiency, incorporating metrics such as finishing, tackling, and positioning. The second principle, completeness, mandates that chosen attributes be available for a sufficient number of players across different teams and years, thereby mitigating sample bias~\cite{groll2015prediction}. Finally, the principle of non-redundancy focuses on minimizing highly correlated attributes or redundant indicators that could introduce multicollinearity into the model. By adhering to these guiding principles we ensured that only features deemed relevant were retained to enhance the robustness and reliability of the subsequent modeling stages.

\subsection{Predictive Modeling Framework}

\subsubsection{Binary Classification Task}
We structured the prediction of match outcomes as a binary classification task, focusing on whether Team~\(A\) wins or Team~\(B\) wins and excluding draws. By centering our approach on classification, we obtain direct probability distributions over possible outcomes with each model yielding a winning probability for both teams that sums to 1.

\subsubsection{Model Families and Rationale}
We selected an different ML algorithms for classification, each offering distinct biases and variance characteristics~\cite{groll2019hybrid}. The models are:
\begin{enumerate}[label=\Roman*.]
    \item A linear approach (Logistic Regression) for interpretability.
    \item Tree-based methods (Random Forest, Gradient Boosting, AdaBoost) known to perform well on tabular data.
    \item \(k\)-Nearest Neighbors to provide additional diversity.
\end{enumerate}

Combining predictions from multiple estimators can enhance accuracy and reduce overfitting risks~\cite{srinivas2022improving}.

\subsubsection{Feature Engineering \& Dimensionality Reduction}
Before training, we standardized aggregated features derived from each team's historical performance statistics. Since each match is defined by two inputs (Team~\(A\) and Team~\(B\)), we created feature representations that captured each team’s relevant attributes. We also applied dimensionality reduction using the Principal Component Analysis (PCA) when beneficial, helping to mitigate high-dimensional noise and potential overfitting~\cite{PCA}.

Following a standard train-test split, we performed k-fold cross-validation (with k=5) on the training set to identify optimal hyperparameters for each model type. For each model type, we selected the hyperparameters that yielded the best accuracy across the k-folds. After identifying the best hyperparameters for each model type, we trained each model with its respective optimal hyperparameters on the entire training set. Finally, we evaluated the selected best models on the held-out test set to ensure robust generalization~\cite{schratz2019hyperparameter}.

\subsection{Inference and Application}

\subsubsection{Prediction via Majority Voting}
Multiple classifiers are involved in the inference (Logistic Regression, Random Forest, XGBoost, AdaBoost, and \(k\)-Nearest Neighbors) \cite{hastie2009elements}, each returning two probabilities (Team~\(A\) win vs. Team~\(B\) win). The winner prediction is determined through majority voting.

\subsubsection{Interpretation and Usage}
The proposed approach for match winner prediction can serve a range of practical and research-related purposes. Coaching staff and analysts may interpret how various features increase win probability, helping to fine-tune tactical decisions. Federation officials and fans can simulate entire tournament brackets by iterating pairwise predictions to gauge likely progression scenarios. Researchers exploring new modeling techniques or additional feature sets can use the approach as a performance baseline for comparative analysis.

\section{Evaluation}

In this section, we evaluate and compare the performance of the proposed ML model against a baseline model using multiple metrics, as summarized in Table~\ref{tab:results_comparison}. Our goal is to highlight both the overall predictive capabilities of each model and the nuances in their performance under different match conditions.

The baseline model was constructed using historical match data from FIFA World Cups dating back to 1930. It relies on the principle that, if two teams have played five or more matches historically, the team with the higher total number of wins is predicted to be victorious. The threshold of five matches was selected based on the 75\textsuperscript{th} percentile of historical head-to-head match counts across all pairs of teams. If fewer than five matches exist in the historical record, the baseline model employs a Weighted Win Ratio (WWR) formula:
\[
\text{WWR} = \left(\frac{v}{v+m}\right) \times R + \left(\frac{m}{v+m}\right) \times C
\]
where $v$ is the total number of matches a team has played, $R$ is its win ratio (computed as total wins divided by $v$), $m$ is a threshold parameter, and $C$ is the overall win ratio computed across all teams.

In contrast, the proposed ML model leverages advanced feature representations derived from the same historical data in order to predict match outcomes. Its design integrates various match-specific features (e.g., team form, past head-to-head performance, etc.) into a supervised learning framework.

Accuracy for both models was calculated based on the standard metric:
\[
\text{Accuracy} = \frac{\text{Number of Correct Predictions}}{\text{Total Number of Predictions}} \times 100\%,
\]
facilitating a straightforward comparison of predictive performance.

\begin{table}[htbp]
    \centering
    \renewcommand{\arraystretch}{1.2} 
    \setlength{\tabcolsep}{10pt} 
    \caption{Performance Comparison Between the Proposed Method and Baseline Model}
    \label{tab:results_comparison}
    \begin{tabular}{lcc}
        \toprule
        \textbf{Evaluation Metric} & \textbf{Proposed Method} & \textbf{Baseline Model} \\
        \midrule
        Overall Accuracy            & \textbf{59.38\%} & 56.25\% \\
        Accuracy (High-scoring)      & 81.25\% & 81.25\% \\
        Accuracy (Low-scoring)       & \textbf{52.08\%} & 47.92\% \\
        \bottomrule
    \end{tabular}
\end{table}

\textbf{Overall Accuracy.} As shown in Table~\ref{tab:results_comparison}, the ML model attains an overall accuracy of 59.38\%, compared to 56.25\% for the baseline. Although the difference of approximately 3\% may appear modest, it signals that the ML model is more effectively exploiting the underlying data patterns, thereby providing a more robust framework for outcome prediction. This improvement suggests that the ML model is able to generalize beyond historical head-to-head records and integrate additional latent factors.

\textbf{Performance in Different Match Scenarios.} In high-scoring matches, both models converge on an identical accuracy of 81.25\%. These matches typically provide clear indicators of in-form attacking teams, which may simplify predictions. However, in low-scoring matches, the ML model exhibits a 4.16\% advantage over the baseline (52.08\% vs. 47.92\%). This highlights its enhanced capability in scenarios where goals are scarce and match outcomes may hinge on subtler factors such as defensive strategy, in-game tactics, or player fitness levels.

\textbf{Predicting Challenging Matches.} Finally, a distinctive strength of the ML model emerges when we isolate matches in which the baseline model failed to predict accurately. Among these 28 so-called \emph{challenging cases}, the ML model successfully predicted 7, thus achieving an accuracy of 25.00\%. Although this figure might seem relatively low, it is notable that these are precisely the matches where simpler, historically informed approaches are most prone to error. The ML model’s success here underscores its ability to capture deeper insights, potentially derived from complex interactions among team attributes, evolving tactics, and situational factors.

\bibliographystyle{IEEEtran}
\bibliography{ ./ref}

\end{document}